\documentclass{article}

\PassOptionsToPackage{numbers,sort&compress}{natbib}
\usepackage[preprint]{neurips_2025}

\usepackage[utf8]{inputenc}
\usepackage[T1]{fontenc}
\usepackage{amsmath, amssymb, amsfonts}
\usepackage{microtype}
\usepackage{graphicx}
\usepackage{booktabs}
\usepackage{array}
\usepackage{colortbl}
\usepackage[hidelinks]{hyperref}
\usepackage{enumitem}
\usepackage{needspace}
\usepackage{url}
\usepackage{adjustbox}
\usepackage{xcolor}

\definecolor{rowhi}{HTML}{FFF8E6}
\definecolor{rowours}{HTML}{E9F6EC}

\title{Self-Distillation Policy Optimization via Visual Feedback: Bridging Code and Visual Artifacts}

\author{%
  Haoyu Dong\\
  Microsoft\\
  \texttt{hadong@microsoft.com}
}

\begin{document}
\maketitle

\begin{abstract}
Code-generating large language models (LLMs) increasingly produce visual artifacts such as charts, web pages, and slides by writing programs that are executed by non-differentiable renderers, committing to code before observing the render. As a result, otherwise executable code often yields artifacts with visually salient defects, including overlapping elements, clipped text, broken alignment, low contrast, and overflow. We study visual-feedback self-distillation for code-generated visual artifacts. We propose Visual-SDPO, a self-distillation policy-optimization framework that treats rendered visual feedback as \emph{privileged context} for a weight-sharing teacher and distills this feedback into a coding student. To make supervision spatially targeted rather than uniform, we introduce \emph{Visual-Grounded Code Credit Weighting}, which traces each detected defect back to the code statements responsible for the affected elements and amplifies the distillation signal on those statements. A sequence-level GRPO (Group Relative Policy Optimization) term complements the dense token-level objective, providing useful training signal even for rollouts whose code fails to execute. We instantiate Visual-SDPO for chart, web/UI, and slide generation with a unified Qwen3-VL-8B-Instruct backbone. Across chart-to-code, UI-to-code, and slide-generation benchmarks (ChartMimic, Design2Code, and AeSlides), Visual-SDPO improves over the zero-shot base by more than 10 absolute points in the primary metric---and over GRPO by at least 2.4 points---with fewer training steps and no added inference-time cost.
\end{abstract}

\section{Introduction}
\label{sec:intro}

Code-driven generation of visual artifacts---charts, slides, and web pages---has become a core capability of general-purpose assistants~\citep{chartmaster, webcode2m, pptagent}. Yet the object optimized by the model (code) and the object evaluated by users or benchmarks (rendered pixels) are separated by a non-differentiable rendering step. The model must commit to code before seeing the render, so syntactically valid programs often produce visibly flawed artifacts: legends overlap data, text is clipped at frame boundaries, elements collide, and shapes become asymmetric.

\begin{figure}[t]
  \centering
  \includegraphics[width=\linewidth]{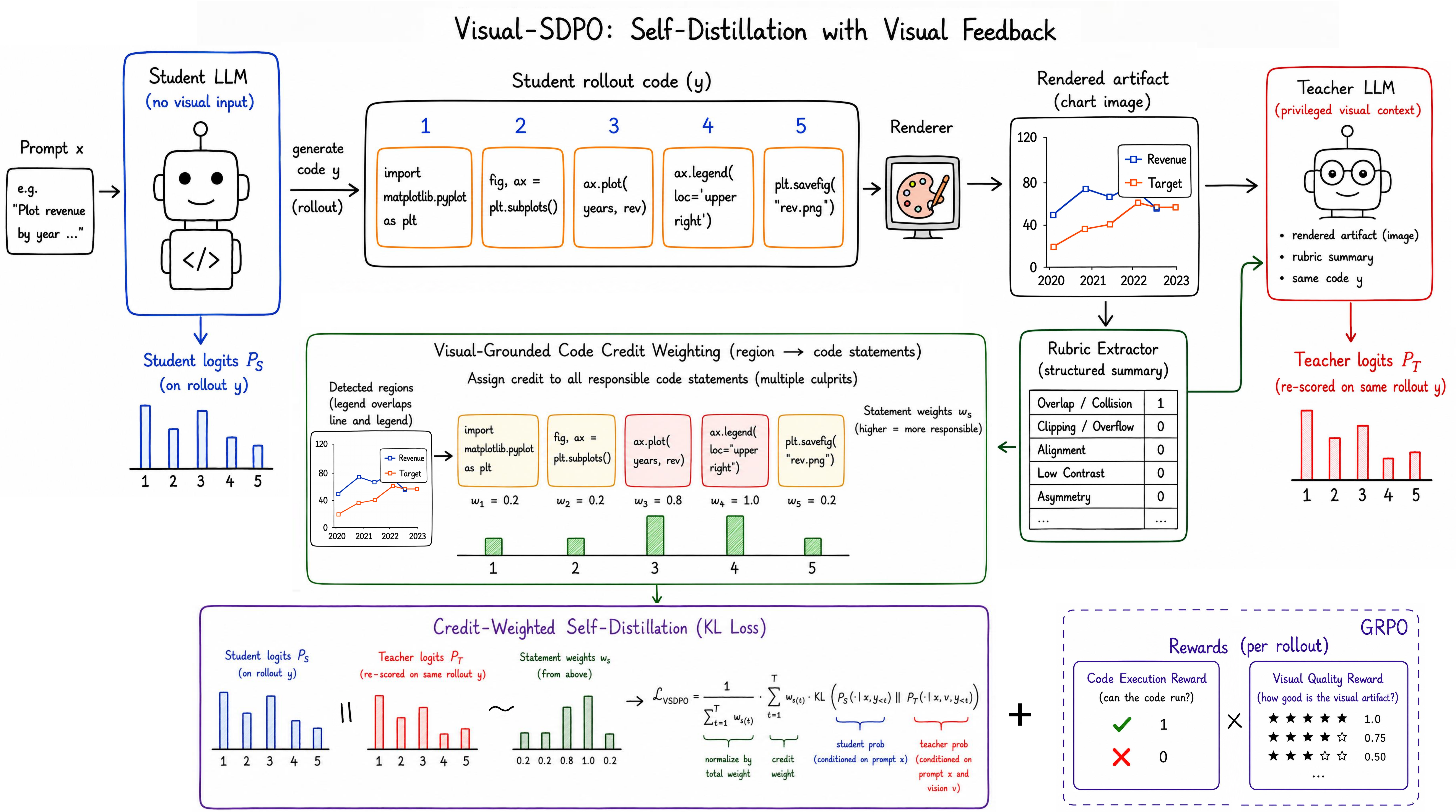}
  \caption{\textbf{Visual-SDPO overview.} The \emph{student LLM} generates a code rollout (shown as a simplified token strip, top center); the \emph{renderer} (matplotlib, Playwright, or python-pptx) produces a rendered artifact. A \emph{defect detector} returns regions around visual problems (legend overlapping a data line in the example shown), and a \emph{rubric extractor} produces a structured binary defect table. Both feed the \emph{teacher LLM}, which shares weights with the student but additionally conditions on this privileged visual context. The \emph{spatial mapping} back-references each defect region to the responsible code statements, producing per-statement responsibility scores $\mathrm{resp}(s)$ that are shared by every token in the statement and converted into per-token weights $w_t = 1 + (\alpha-1)\cdot\mathrm{resp}(s(t))$, normalized. These weights scale a token-level KL divergence between student and teacher distributions, concentrating the distillation gradient on the code statements that produced the defective regions. A sequence-level GRPO reward (bottom right) combines with this distillation loss to form the full training objective. At inference, the student operates without the renderer, defect detector, rubric extractor, or teacher pass.}
  \label{fig:overview}
\end{figure}

Existing methods address this code--vision gap in two main ways, both with important limitations. \emph{Inference-time visual reflection}~\citep{pptagent, posteragent} runs render--critique--revise loops until a VLM critic accepts the artifact. Each iteration re-executes the renderer, re-queries the critic, and regenerates code, so deployment cost grows with the number of polishing rounds and requires both a renderer and a VLM critic at inference. \emph{Visual-reward RL}~\citep{aeslides, chartmaster, msrl, rrvf, rlrf, ui2codeN} instead optimizes GRPO or DPO against an image-level scalar reward. This supervision is sparse and trajectory-level: per-element diagnostics are collapsed into a single number before they can shape the gradient.

We propose a third path, summarized in Figure~\ref{fig:overview}: use rendered visual feedback as \emph{privileged context} for self-distillation~\citep{sdpo}. The teacher is conditioned on the rendered artifact, or a structured summary of it, whereas the student is conditioned only on the original prompt. Token-level distillation then transfers visual understanding into a student that requires no renderer at inference. Unlike a scalar reward, this feedback is \emph{localizable}: it identifies which element overlaps, which text is clipped, and which color combination is unreadable.

Uniform token-level distillation, however, can induce policy drift by matching teacher logits at positions unrelated to the visual outcome---control-flow tokens, variable declarations, and other non-visual scaffolding. To mitigate this, we route the distillation gradient toward the code statements responsible for rendered defects, while retaining a baseline gradient on non-defect statements. We formalize this mechanism as \emph{Visual-Grounded Code Credit Weighting}. We pair the token-level objective with a sequence-level GRPO reward because the two signals are complementary: distillation provides dense, localizable supervision, while the trajectory-level reward anchors training and transfers across domains. Together, they make training both stable and sample-efficient. Visual-SDPO lies at the intersection of privileged self-distillation, visual-feedback training, and code-generated visual artifacts.

\paragraph{Contributions.}
\begin{enumerate}[leftmargin=*, topsep=2pt, itemsep=1pt]
  \item Visual-SDPO, a visual-feedback self-distillation framework for code-generated visual artifacts, in which a weight-sharing teacher observes the rendered artifact as privileged context and distills that feedback into a coding student.
  \item Visual-Grounded Code Credit Weighting, a visual-grounded code credit-assignment mechanism that traces rendered visual defects back to the responsible code statements and uses them to weight token-level distillation. We combine this token-level signal with a complementary sequence-level GRPO reward.
  \item A unified instantiation and evaluation across chart-to-code, UI-to-code, and slide-generation benchmarks---ChartMimic, Design2Code, and AeSlides---on a single Qwen3-VL-8B-Instruct backbone, improving over the zero-shot base by more than 10 absolute points in the primary metric.
\end{enumerate}

\section{Method}
\label{sec:method}

\subsection{Problem Setup}
Let $x$ denote a task input---a natural-language instruction, optionally paired with a reference image (e.g., a screenshot to reproduce). The policy $\pi_\theta$ generates a code sequence $y = (y_1, \dots, y_T)$. A deterministic renderer $R$ maps $y$ to a rendered artifact $R(y)$, such as an HTML page, a chart image, or a slide deck. A visual diagnostic extractor converts the rendered artifact into privileged visual context $v$: either the rendered image itself or a rubric-structured score vector over defect categories such as alignment, overlap, contrast, and overflow. From the same rendered artifact, a defect detector returns a set $D = \{(\mathrm{region}_i, \mathrm{severity}_i)\}$ of detected visual defects, each with a binary severity label. A region-to-code mapping $M$ identifies, for each statement $s$ in $y$---a Python source line for charts, an HTML element for web pages, or a python-pptx shape-creating call for slides---the rendered regions produced by that statement.

\subsection{Visual-Feedback Self-Distillation}
The student takes only the original input,
\begin{equation}
  p_S(y_t \mid y_{<t}, x) = \pi_\theta(y_t \mid y_{<t}, x).
\end{equation}
The teacher shares weights with the student but additionally conditions on the visual diagnostic produced from the student's own rollout:
\begin{equation}
  p_T(y_t \mid y_{<t}, x, v) = \pi_\theta(y_t \mid y_{<t}, x, v),
\end{equation}
where $v$ is either the raw rendered image or a structured rubric vector. Crucially, the teacher does not regenerate $y$; it re-scores the student's tokens under privileged context. At each update, $p_T$ is computed by a second forward pass of the current policy and is detached before the KL is taken, so gradients flow only through the student and no separate teacher parameters are introduced. The Visual-Feedback Self-Distillation loss is the token-level KL between the student and teacher distributions on the student's rollout:
\begin{equation}
  \mathcal{L}_{\mathrm{VFD}}(\theta) = \mathbb{E}_{x,\, y \sim \pi_\theta(\cdot \mid x)} \!\left[\, \sum_{t=1}^{T} \mathrm{KL}\!\big(p_S(\cdot \mid y_{<t}, x) \,\big\|\, p_T(\cdot \mid y_{<t}, x, v)\big) \,\right].
\end{equation}
An importance-weighted variant corrects for distribution mismatch between the teacher and student conditioning contexts; we adopt this stabilized variant in practice.

\paragraph{Two channels for the visual diagnostic.}
We instantiate the visual signal $v$ in two complementary forms. \textbf{Image channel.} When the policy is an image-conditioned VLM (Qwen3-VL-8B), $v$ is the rendered image. The teacher receives the prompt together with a screenshot of the artifact rendered from the student's code, whereas the student receives only the prompt, or the prompt plus an input reference image depending on the task. If the student's code fails to execute and no rendered artifact is produced, we replace the image channel with the Python or browser execution error message---a compile error, runtime exception, or traceback---which the teacher consumes in place of the screenshot. This recovers the original SDPO formulation of textual error feedback as privileged context. \textbf{Rubric channel.} When the policy is text-only, we apply a lightweight pretrained \emph{visual rubric extractor} to the rendered artifact and obtain a structured score vector over a domain-appropriate set of defect categories. For each domain, this schema matches the verifiable axis set used by the GRPO reward (Section~\ref{sec:objective}): aspect ratio, whitespace, collision, and imbalance for slides; CLIP similarity, block-match, text, position, and color for web pages; and text, layout, chart type, color, execution validity, and visual similarity for charts. The teacher receives $v$ as serialized JSON-like text appended to the prompt. These signals are complementary by construction: the rubric and the detected defect regions re-express, in a structured and spatially localized form, the same quality dimensions that the verifiable GRPO reward already measures, whereas the rendered screenshot lets the teacher's visual understanding supply holistic, fine-grained feedback that no fixed rubric or scalar reward enumerates.

\subsection{Visual-Grounded Code Credit Weighting}
Vanilla token-level KD treats every output token equally:
$\mathcal{L}_{\mathrm{KD}} = (1/T) \sum_t \mathrm{KL}(p_S(\cdot \mid x, y_{<t}) \,\|\, p_T(\cdot \mid x, v, y_{<t}))$.
This uniformity distracts the gradient in code-to-artifact generation, where most output tokens---variable names, control flow, container scaffolding---do not generate any visible artifact element~\citep{antisd}. Tokens that \emph{do} generate elements involved in detected visual defects should receive stronger supervision, because they are the positions where a gradient update is most likely to improve the rendered artifact. Visual-Grounded Code Credit Weighting operationalizes this intuition through three components: defect detection, region-to-code mapping, and a statement-weighted KL loss.

\paragraph{Defect detection.}
From the rendered artifact $R(y)$, we obtain a set of detected visual defects $D = \{(\mathrm{region}_i, \mathrm{severity}_i)\}$, where each defect consists of a region on the rendered canvas and a binary severity label $\mathrm{severity}_i \in \{0,1\}$. The detector's regions feed the region-to-code mapping; in parallel, a \emph{rubric extractor} produces a structured binary defect table over a domain-specific set of axes, which is appended to the teacher's privileged context. We use the \emph{same} per-domain axis sets for both the rubric channel and the GRPO visual reward (Section~\ref{sec:objective}), coupling the two signals by construction:
\begin{itemize}[leftmargin=*, topsep=2pt, itemsep=1pt]
  \item \textbf{Web/UI (Design2Code).} Five axes inherited from the Design2Code evaluation suite: CLIP similarity, block-match, text, position, color.
  \item \textbf{Slide (AeSlides).} Four verifiable axes from AeSlides: aspect ratio, whitespace, collision, imbalance.
  \item \textbf{Chart (ChartMimic).} Execution validity and four low-level format axes---text, layout, chart type, and color---together with a visual-similarity score against the reference image.
\end{itemize}
The rubric extractor returns a binary value for each axis. The GRPO term in Eq.~\eqref{eq:rvis} then takes a weighted sum of the per-axis \emph{satisfaction} indicators (each $1$ when the corresponding axis is satisfied), with non-negative weights summing to one, to form the scalar visual reward $r_{\mathrm{vis}}(y) \in [0,1]$. Thus, the same visual evidence drives both the teacher's privileged context (a per-axis structured table) and the sequence-level reward (a per-axis weighted sum), without introducing additional supervision channels.

\paragraph{Region-to-code mapping.}
For each detected defect region, we identify the \emph{code statements} that produced the element(s) overlapping that region. We operate at statement granularity rather than token granularity because each visible artifact element is typically created by a coherent \emph{act of generation} in the student's code---a matplotlib API call, an HTML element, or a python-pptx shape constructor---that spans multiple semantically coupled tokens. We use a mapping $M$ that returns, for each statement $s$, the set of rendered regions it produced; it can be obtained in two ways.

\textbf{Runtime introspection.} We instrument the renderer to record, for each rendered element, the source statement that emitted it. For matplotlib, artist constructors are hooked to capture the source line in the user frame via \texttt{sys.\_getframe()} and the rendered region via \texttt{get\_window\_extent()}; for web pages, HTML elements carry \texttt{data-src-*} attributes that persist in the DOM and are read together with \texttt{getBoundingClientRect()}; for slides, python-pptx shape constructors are hooked analogously. The instrumentation does not alter the prompt, generated code, or rendered artifact observed by the student.

\textbf{VLM-based mapping.} Alternatively, the same Qwen3-VL teacher used for privileged-context distillation can serve as the mapping module. Given the student's code and the rendered image annotated with defect regions, the VLM identifies which statement is responsible for each defect. This route is renderer-agnostic and readily portable to new toolchains, at the cost of additional teacher forward passes. In our experiments, we use runtime introspection when instrumentation is available and fall back to VLM-based mapping for toolchains we have not instrumented.

A single detected defect may trace back to multiple statements. The formulation below handles this naturally: each statement's responsibility is computed independently against every defect region, so one defect can flag several statements, and one statement can be flagged by several co-located defects.

\paragraph{Statement-weighted KL loss.}
Let $S$ denote the set of statements in the student's rollout. For each statement $s \in S$, the mapping $M$ returns the set of rendered-element regions that statement produced. We compute a per-statement responsibility score against detected defects $D$ and convert it into a per-statement weight via a single amplification factor $\alpha \ge 1$:
\begin{align}
  \mathrm{resp}(s) &= \max_{i \in D} \Big[\, \big(\max_{b \in M(s)} \mathrm{IoU}(b, \mathrm{region}_i)\big) \cdot \mathrm{severity}_i \,\Big], \\
  w_s &= 1 + (\alpha - 1) \cdot \mathrm{resp}(s) \;\in\; [1, \alpha].
\end{align}
Here $b$ ranges over the rendered regions in $M(s)$, so a multi-element statement is scored by its best-overlapping region. We assign each rollout token $t$ to the smallest enclosing parsed statement $s(t)$, which inherits that statement's weight, $w_t = w_{s(t)}$; whitespace and separator tokens inherit the nearest preceding statement, and any token without a parsed statement receives weight $1$. The full Visual-SDPO loss is the statement-weighted per-token KL:
\begin{equation}\label{eq:vsdpo}
  \mathcal{L}_{\mathrm{VSDPO}}(\theta) = \frac{1}{\sum_t w_t} \sum_t w_t \cdot \mathrm{KL}\!\big(p_S(\cdot \mid x, y_{<t}) \,\big\|\, p_T(\cdot \mid x, v, y_{<t})\big).
\end{equation}
Tokens in statements that produce no rendered element (imports, declarations, control flow) receive $\mathrm{resp}(s) = 0$ and weight $1$. Tokens in a statement whose element lies fully inside a detected defect region receive weight $\alpha$; because severity is binary, the smooth interpolation between $1$ and $\alpha$ comes entirely from the IoU overlap. Normalizing by $\sum_t w_t$ keeps the loss scale comparable to uniform KD, so $\alpha$ can be tuned independently of the average defect density; setting $\alpha = 1$ recovers uniform visual SDPO.

\subsection{Final Training Objective}
\label{sec:objective}
We optimize token-level distillation and the sequence-level GRPO reward jointly because they provide complementary supervision. The distillation term supplies a dense, sample-efficient, and \emph{localizable} signal, allowing credit weighting to concentrate updates on statements responsible for visual defects. The GRPO term contributes an execution-grounded outcome reward that anchors training across heterogeneous toolchains. Optimizing their sum lets each signal compensate for the other's blind spots:
\begin{equation}\label{eq:final}
  \mathcal{L}(\theta) = \mathcal{L}_{\mathrm{VSDPO}}(\theta) + \beta \cdot \mathcal{L}_{\mathrm{GRPO}}(\theta).
\end{equation}
The hyperparameter $\beta$ trades off the dense per-token distillation signal ($\mathcal{L}_{\mathrm{VSDPO}}$, driven by rendered visual feedback and credit-weighted by $\alpha$) against the sparse sequence-level reward ($\mathcal{L}_{\mathrm{GRPO}}$). The GRPO term uses a multiplicative composite reward,
\begin{equation}\label{eq:rvis}
  r(y) = r_{\mathrm{exec}}(y) \cdot r_{\mathrm{vis}}(y),
\end{equation}
where $r_{\mathrm{exec}}(y) \in \{0, 1\}$ is the binary execution-success indicator (1 if the rollout's code executes and produces a rendered artifact, 0 otherwise), and $r_{\mathrm{vis}}(y) \in [0, 1]$ is a graded visual-quality score derived from the rubric extractor (a count-based aggregate of the structured binary defect table) or, optionally, an external VLM judge; although each axis is binary, their weighted aggregate is continuous in $[0,1]$. The multiplicative form assigns zero reward to execution failures and gives positive reward only when the code executes and renders an acceptable artifact.

On failed rollouts ($r_{\mathrm{exec}}(y) = 0$, hence $r(y) = 0$), we set $r_{\mathrm{vis}}(y) = 0$ by convention and do not evaluate the extractor; we additionally pass the Python or browser error trace to the teacher as textual privileged context, recovering the original SDPO formulation so that $\mathcal{L}_{\mathrm{VSDPO}}$ still contributes useful supervision. On successful rollouts, the teacher receives the rendered image and the structured rubric as visual privileged context, while the GRPO term rewards executability and visual quality.

\textit{Optional warm-start.} For domains with a labeled reference corpus (e.g., Chart2Code-160K), we additionally apply a standard supervised next-token loss $\mathcal{L}_{\mathrm{SFT}}$ against the reference code during an initial warm-start phase only, before the Visual-SDPO + GRPO joint training.

\section{Experiments}
\label{sec:experiments}

\subsection{Datasets}
\textbf{Training corpora.} We train separately in each domain and never mix corpora across domains.
\begin{itemize}[leftmargin=*, topsep=2pt, itemsep=1pt]
  \item \textbf{Chart.} Chart2Code-160K~\citep{chartcoder}, filtered to its executable subset.
  \item \textbf{Web/UI.} WebCode2M~\citep{webcode2m} + WebSight~\citep{websight}.
  \item \textbf{Slide (instruction-only).} AeSlides-7k page-level prompts (used by AeSlides~\citep{aeslides} as the verifiable-reward RL corpus).
\end{itemize}

\textbf{Evaluation benchmarks.} We evaluate on ChartMimic~\citep{chartmimic} (Direct Mimic), Design2Code~\citep{design2code}, and the AeSlides verifiable metrics (aspect ratio, whitespace, collision, imbalance) on the AeSlides-7k-eval split.

\subsection{Baselines}
Our experimental design controls the backbone and the per-domain training corpus across methods: within each result table, all baselines start from the same Qwen3-VL-8B-Instruct backbone and use the same per-domain training corpus.
\needspace{9\baselineskip}
\begin{itemize}[leftmargin=*, topsep=2pt, itemsep=1pt]
  \item \textbf{SFT.} Standard supervised fine-tuning on the per-domain training corpora.
  \item \textbf{OPSD (ref-code privileged teacher).} Self-distillation where the teacher has access to the ground-truth reference code (where available) rather than the rendered artifact. This isolates reference-code privilege from visual privilege.
  \item \textbf{GRPO with visual reward.} This baseline renders rollouts, computes a scalar visual reward from a domain-appropriate metric, and updates the policy with GRPO.
\end{itemize}

\subsection{Main Results --- Chart Domain (Table~\ref{tab:chartmimic})}

\begin{table}[t]
  \centering
  \small
  \caption{ChartMimic Direct Mimic results. Exec = execution success rate; Low = average of the four low-level F1 axes (Text, Layout, Chart-Type, Color); High = GPT-4o judge score; Overall = (Low $+$ High)$/2$. All Visual-SDPO ablations (highlighted) use a unified Qwen3-VL-8B-Instruct backbone. $\Delta$ = absolute Overall improvement over the zero-shot Qwen3-VL-8B-Instruct base.}
  \label{tab:chartmimic}
  \adjustbox{max width=\linewidth}{
  \begin{tabular}{lcccccc}
    \toprule
    \textbf{Method} & \textbf{Backbone} & \textbf{Exec} & \textbf{Low} & \textbf{High} & \textbf{Overall $\uparrow$} & $\Delta$ \\
    \midrule
    Qwen3-VL-8B-Instruct (zero-shot) & Qwen3-VL-8B & 81.7 & 62.9 & 72.9 & 67.9 & --- \\
    \rowcolor{rowhi} \emph{+ SFT} (Chart2Code-160K) & Qwen3-VL-8B & 90.3 & 69.9 & 77.2 & 73.6 & +5.7 \\
    \rowcolor{rowhi} \emph{+ OPSD} (ref-code priv. teacher) & Qwen3-VL-8B & 92.3 & 73.9 & 80.1 & 77.0 & +9.1 \\
    \rowcolor{rowhi} \emph{+ GRPO} (visual-similarity reward) & Qwen3-VL-8B & 92.9 & 72.6 & 79.7 & 76.2 & +8.3 \\
    \rowcolor{rowours} \textbf{+ Visual-SDPO (ours, full)} & Qwen3-VL-8B & \textbf{92.7} & \textbf{74.8} & \textbf{82.3} & \textbf{78.6} & \textbf{+10.7} \\
    \bottomrule
  \end{tabular}}
\end{table}

\paragraph{Ablation analysis.}
The four-step ablation isolates the contribution of each supervision component and reveals a notable non-monotone pattern on ChartMimic. (1) \emph{+ SFT} on Chart2Code-160K lifts Overall from 67.9 to 73.6 ($+5.7$), with most of the gain coming from the Low-level axes (62.9 $\to$ 69.9). (2) \emph{+ OPSD} ($+3.4$ over SFT, $+9.1$ over base) adds a privileged-context teacher conditioned on reference code and yields the largest single-step gain over SFT among the privileged-context and RL stages. Low-level F1 rises sharply (69.9 $\to$ 73.9), suggesting that matplotlib code benefits strongly from token-level structural disambiguation---axis APIs, color and marker spellings, and legend positions---that a sequence-level visual reward cannot provide. (3) \emph{+ GRPO} with a sequence-level visual-similarity reward reaches 76.2 Overall ($+8.3$ over base, 0.8 below OPSD), staying close to OPSD on the High-level GPT-4o judge score (79.7 vs.\ 80.1) but trailing further on Low-level F1 (72.6 vs.\ 73.9). On chart-to-code, dense token-level supervision from reference code thus appears more informative than a scalar trajectory-level visual reward. (4) \emph{+ Visual-SDPO} ($+10.7$ over base, $+1.6$ over the stronger single-signal baseline) indicates that the two signals are complementary: dense per-statement credit plus sequence-level reward outperforms either signal alone on both Low-level F1 (74.8) and High-level score (82.3).

\subsection{Main Results --- Web/UI Domain (Table~\ref{tab:design2code})}

\begin{table}[t]
  \centering
  \small
  \caption{Design2Code results on the public eval set. Overall is the arithmetic mean of the five low-level axes. All Visual-SDPO ablations (highlighted) use Qwen3-VL-8B-Instruct.}
  \label{tab:design2code}
  \adjustbox{max width=\linewidth}{
  \begin{tabular}{lcccccccc}
    \toprule
    \textbf{Method} & \textbf{Backbone} & \textbf{CLIP} & \textbf{Block} & \textbf{Text} & \textbf{Position} & \textbf{Color} & \textbf{Overall $\uparrow$} & $\Delta$ \\
    \midrule
    Qwen3-VL-8B-Instruct (zero-shot) & Qwen3-VL-8B & 85.4 & 50.2 & 78.1 & 71.9 & 74.7 & 72.1 & --- \\
    \rowcolor{rowhi} \emph{+ SFT} (WebSight + WebCode2M) & Qwen3-VL-8B & 87.1 & 59.6 & 82.3 & 77.2 & 78.8 & 77.0 & +4.9 \\
    \rowcolor{rowhi} \emph{+ OPSD} (ref-code priv. teacher) & Qwen3-VL-8B & 87.4 & 62.3 & 82.9 & 78.4 & 81.8 & 78.6 & +6.5 \\
    \rowcolor{rowhi} \emph{+ GRPO} (5-axis weighted reward) & Qwen3-VL-8B & 88.3 & 64.5 & 83.3 & 78.7 & 85.1 & 80.0 & +7.9 \\
    \rowcolor{rowours} \textbf{+ Visual-SDPO (ours, full)} & Qwen3-VL-8B & \textbf{89.2} & \textbf{69.7} & \textbf{84.9} & \textbf{82.4} & \textbf{86.8} & \textbf{82.6} & \textbf{+10.5} \\
    \bottomrule
  \end{tabular}}
\end{table}

\paragraph{Ablation analysis.}
Design2Code exhibits a different per-axis profile from ChartMimic. (1) \emph{+ SFT} on WebSight + WebCode2M ($+4.9$ Overall) improves Block-Match the most (50.2 $\to$ 59.6), indicating that the model learns to emit a more faithful container hierarchy. The CLIP gain is smaller (85.4 $\to$ 87.1), consistent with the base Qwen3-VL model already being perceptually strong. (2) \emph{+ OPSD} ($+1.6$ over SFT) primarily improves Color (78.8 $\to$ 81.8), consistent with reference-HTML conditioning helping to resolve CSS color-naming ambiguities. (3) \emph{+ GRPO} with the standard Design2Code 5-axis weighted reward ($+1.4$ over OPSD) further improves Color (81.8 $\to$ 85.1) and Block-Match (62.3 $\to$ 64.5), with only a marginal gain on Text (82.9 $\to$ 83.3). (4) \emph{+ Visual-SDPO} ($+2.6$ over GRPO, $+10.5$ over base) yields its largest gains on the spatial axes---Position (78.7 $\to$ 82.4) and Block-Match (64.5 $\to$ 69.7)---where region-to-code mapping over the HTML DOM should be most informative: an overlapping element can be traced back to the specific \texttt{<div>} whose CSS width or position properties need to change.

\subsection{Main Results --- Slide Domain (Table~\ref{tab:aeslides})}

\begin{table}[t]
  \centering
  \small
  \caption{AeSlides results on the AeSlides-7k-eval split (with decks held out). Avg is the arithmetic mean of the four AeSlides verifiable axes (aspect-ratio compliance, whitespace, collision, imbalance), normalized so that higher = better quality. All ablations (highlighted) use Qwen3-VL-8B-Instruct. OPSD is omitted because the instruction-only slide corpus has no reference code for the privileged teacher.}
  \label{tab:aeslides}
  \begin{tabular}{lccc}
    \toprule
    \textbf{Method} & \textbf{Backbone} & \textbf{Avg $\uparrow$} & $\Delta$ \\
    \midrule
    Qwen3-VL-8B-Instruct (zero-shot) & Qwen3-VL-8B & 49.5 & --- \\
    \rowcolor{rowhi} \emph{+ SFT} (AeSlides-7k-train) & Qwen3-VL-8B & 52.8 & +3.3 \\
    \rowcolor{rowhi} \emph{+ GRPO} (AeSlides 4-axis reward) & Qwen3-VL-8B & 58.2 & +8.7 \\
    \rowcolor{rowours} \textbf{+ Visual-SDPO (ours, full)} & Qwen3-VL-8B & \textbf{60.7} & \textbf{+11.2} \\
    \bottomrule
  \end{tabular}
\end{table}

\paragraph{Ablation analysis.}
On AeSlides, the largest single-stage gain comes from the verifiable-reward (GRPO) stage; OPSD is omitted here because the instruction-only slide corpus provides no reference code for the privileged teacher. (1) \emph{+ SFT} on AeSlides-7k-train ($+3.3$ Avg) gives the smallest SFT gain across domains: it teaches general slide-page structure but does not directly optimize the four rule-based metrics that define the evaluation. (2) \emph{+ GRPO} with the 4-axis verifiable reward ($+5.4$ over SFT) is the largest single-step contributor, consistent with the reward coinciding with the evaluation metric, which lets sequence-level RL directly optimize the four verifiable axes. (3) \emph{+ Visual-SDPO} ($+2.5$ over GRPO, $+11.2$ over base) suggests that token-level credit assignment still adds value on top of metric-aligned GRPO: per-statement responsibility ties each aspect-ratio violation back to the responsible shape/layout statement, rather than rewarding only the whole rollout.

\paragraph{Training efficiency.} The dense per-token distillation signal also makes Visual-SDPO more sample-efficient than sequence-level RL: it matches the performance GRPO attains at full training with far fewer rollout generations and renderer evaluations---on average only about 29\% of GRPO's rollout budget across the three domains---and correspondingly less wall-clock training.

\section{Related Work}
\label{sec:related}

\subsection{Self-Distillation with Privileged Information}
Visual-SDPO extends on-policy self-distillation and privileged-information distillation from textual reasoning or code feedback to rendered visual feedback. Self-distillation methods train a student and a teacher that share weights but receive different input contexts, then minimize a token-level divergence between their distributions on the student's own rollouts. The privileged-information view of distillation traces back to Lopez-Paz et al.~\citep{lopezpaz2016}, who unified knowledge distillation with Vapnik's privileged-information framework, and was extended to on-policy language-model distillation by Agarwal et al.~\citep{gkd}, who proposed learning from the student's own generations rather than from a fixed dataset. Recent work explores several forms of privileged context: ground-truth answers or reference reasoning traces in math (OPSD~\citep{opsd}, HDPO~\citep{hdpo}); environment error feedback or successful sibling rollouts in code (SDPO~\citep{sdpo}); and consensus- or uncertainty-aware gating of the divergence signal~\citep{gates, egrsd}. The broader literature on on-policy distillation~\citep{opsdsurvey} further documents this design space, while a complementary anti-self-distillation line~\citep{antisd} cautions against token-uniform KD on positions that the privileged signal does not identify. We extend this family in two orthogonal directions: (i) the privileged channel becomes a \emph{visual} signal---either the rendered artifact itself or a rubric-structured visual summary---and (ii) we depart from uniform token weighting by introducing \emph{Visual-Grounded Code Credit Weighting}, which maps detected visual defects back to the specific code statements responsible for them.

\subsection{Visual-Feedback Training for Code Generation}
Visual-SDPO differs from visual-reward RL methods for chart-to-code, UI-to-code, and slide generation by using rendered feedback as both privileged teacher context and a token-level code-credit signal. A growing body of work exploits the rendered visual surface of generated code as a training signal. For chart code generation, MSRL~\citep{msrl} and ChartMaster~\citep{chartmaster} optimize chart-to-code generation with chart-similarity rewards, while ChartEditor~\citep{charteditor} studies RL for robust chart editing; RRVF~\citep{rrvf} learns from rendered images alone; and RLRF~\citep{rlrf} applies rendering-aware RL to vector graphics. For web/UI generation, UI2Code$^{\mathrm{N}}$~\citep{ui2codeN} frames UI-to-code as test-time scalable interactive visual optimization, VinciCoder~\citep{vincicoder} uses coarse-to-fine visual RL, ReLook~\citep{relook} grounds RL in a multimodal LLM critic, ScreenCoder~\citep{screencoder} modularizes frontend generation with multimodal agents, and DesignCoder~\citep{designcoder} adds hierarchy-aware self-correction. For slides, AeSlides~\citep{aeslides} uses verifiable aesthetic rewards over rendered slide pages. The data side is supported by WebSight~\citep{websight}, WebCode2M~\citep{webcode2m}, and Web2Code~\citep{web2code}, with Pix2Struct~\citep{pix2struct} and Pix2Code~\citep{pix2code} as foundational precedents; inference-time systems such as DCGen~\citep{dcgen}, PPTAgent~\citep{pptagent}, and Paper2Poster~\citep{posteragent} further demonstrate the importance of execution, rendering, or evaluation loops in visual artifact generation. Along two axes, existing methods can be classified by whether the visual signal is a \emph{scalar reward} or a \emph{structured diagnostic}, and whether the critic is the \emph{same model} as the policy or an \emph{external VLM}. Among the training methods above, many reduce rendered feedback to scalar image-level rewards and optimize GRPO or DPO with an external VLM, learned judge, or embedding-based metric; even methods that expose rubric categories typically aggregate them back to a per-rollout scalar.

The visual signal must ultimately be attributed to the responsible code. Mapping rendered output back to its source has a long history outside LLM training: program slicing~\citep{weiser1981program} and Whyline~\citep{whyline} surface responsible statements for human debugging; the DWARF debug format~\citep{dwarf5} maps program counters to source lines; browser DevTools links DOM nodes to HTML/CSS source positions; RenderDoc and Nsight Shader Debugger~\citep{renderdoc, nsight} attribute rendered pixels to draw calls and shader source lines; Hoffswell et al.'s reactive Vega debugger~\citep{hoffswell2016} links chart marks to spec elements via the Vega-Lite scenegraph; and PyTorch FX~\citep{pytorchfx2022} preserves stack-trace metadata on IR nodes. These systems expose mappings at \emph{inspection time} for human debugging or workflow auditing. Visual-Grounded Code Credit Weighting differs by computing the mapping during model rollout and consuming it as a dense per-statement training signal that scales token-level KD weights. To our knowledge, prior work has not explicitly combined output-to-source mapping, structured visual diagnostics, and privileged self-distillation as a per-statement learning signal for code-generating LLMs that produce rendered visual artifacts.

\section{Conclusion}
\label{sec:conclusion}

We presented Visual-SDPO, a self-distillation framework that uses rendered visual feedback as privileged context for a weight-sharing teacher, together with \emph{Visual-Grounded Code Credit Weighting}, which traces detected visual defects back to the responsible code statements and amplifies the per-token distillation signal there, and a sequence-level GRPO term that complements this dense token-level objective. On a unified Qwen3-VL-8B-Instruct backbone, Visual-SDPO improves over the zero-shot base by more than 10 absolute points in the primary metric across chart-to-code, UI-to-code, and slide-generation benchmarks (ChartMimic, Design2Code, and AeSlides), and over GRPO by at least 2.4 points. Natural extensions include scaling the training corpus to a chart-RL regime comparable to recent large-scale visual-reward work, investigating differentiable rendering as a third visual-feedback channel, and extending region-to-code mapping from static artifacts to multi-step interactive UI flows.

\bibliographystyle{plainnat}
\bibliography{references}

\end{document}